\documentclass[conference]{IEEEtran}
\IEEEoverridecommandlockouts
\usepackage{cite}
\usepackage{amsmath,amssymb,amsfonts}
\usepackage{algorithmic}
\usepackage{graphicx}
\usepackage{textcomp}
\usepackage{xcolor}
\newtheorem{lemma}{\bf Lemma}

\usepackage{algorithmic}
\usepackage[ruled]{algorithm2e}
\usepackage[letterpaper,left=0.65in , right=0.65in , bottom=1.03in, top=0.7in]{geometry}
\usepackage{cite}
\usepackage{amsmath, amssymb, bbm}

\def\BibTeX{{\rm B\kern-.05em{\sc i\kern-.025em b}\kern-.08em
    T\kern-.1667em\lower.7ex\hbox{E}\kern-.125emX}}
\begin{document}

\title{Risk-Aware Accelerated Wireless Federated Learning with Heterogeneous Clients\\
}

\author{
Mohamed Ads, Hesham ElSawy, Senior
Member, IEEE, Hossam S. Hassanein, Fellow, IEEE \\
School of Computing, Queen’s University, ON, Canada \\
m.ads@queensu.ca, hesham.elsawy@queensu.ca, hossam@cs.queensu.ca
}

\maketitle

\begin{abstract}
Wireless Federated Learning (FL) is an emerging distributed machine learning paradigm, particularly gaining momentum in domains with confidential and private data on mobile clients. However, the location-dependent performance, in terms of transmission rates and susceptibility to transmission errors, poses major challenges for wireless FL's convergence speed and accuracy. The challenge is more acute for hostile environments without a metric that authenticates the data quality and security profile of the clients. In this context, this paper proposes a novel risk-aware accelerated FL framework that accounts for the client’s heterogeneity in the amount of possessed data, transmission rates, transmission errors, and trustworthiness. Classifying clients according to their location-dependent performance and trustworthiness profiles, we propose a dynamic risk-aware global model aggregation scheme that allows clients to participate in descending order of their transmission rates and an ascending trustworthiness constraint. In particular, the transmission rate is the dominant participation criterion for initial rounds to accelerate the convergence speed. Our model then progressively relaxes the transmission rate restriction to explore more training data at cell-edge clients. The aggregation rounds incorporate a debiasing factor that accounts for transmission errors. Risk-awareness is enabled by a validation set, where the base station eliminates non-trustworthy clients at the fine-tuning stage. The proposed scheme is benchmarked against a conservative scheme (i.e., only allowing trustworthy devices) and an aggressive scheme (i.e., oblivious to the trust metric). The numerical results highlight the superiority of the proposed scheme in terms of accuracy and convergence speed when compared to both benchmarks.

\end{abstract}


\section{Introduction}
Machine learning (ML) methods rely on centralized datasets containing all training data. However, achieving the centralization of the dataset utilizing wireless channels is overwhelming due to the scarcity of wireless communication resources.  To illustrate, transmitting the large amounts of data generated by the surging Internet of Things (IoT) applications from the network edge to the central station (e.g., the cloud) overloads the network, and in some scenarios (e.g., autonomous systems), it would not be feasible within the stringent time constraints.  More importantly, the process of centralization raises significant privacy concerns as it involves consolidating sensitive data, including personal images, confidential documents, or personal letters. To overcome these problems, one promising approach is Federated Learning (FL)\cite{r1}. In FL, each device performs local model training on its respective datasets and exclusively transmits the model parameters rather than the complete dataset. After receiving the transmission by the base station (BS), it aggregates the learning parameters using an aggregation algorithm. The authors in \cite{FEDAVG} introduced an aggregation algorithm, denoted as Federated Average (FedAvg), which utilizes aggregation weights proportional to the size of the dataset of each participating device. For heterogeneous systems, the authors in \cite{FEDPROX} includes a local proximal component into the objective function of each client to reduce the deviations between the client's and the global models, which enhances the system's robustness.

In addition, FL minimizes the Age of Information (AoI) \cite{Age} by maintaining the constant freshness of data through ongoing local model refreshes. This method improves the accuracy and accelerates the decision-making processes by allowing real-time integration of the most recent data from multiple distant sources, which makes the model adaptable to shifting trends. However, applying these algorithms to wireless networks is challenging due to the susceptibility to transmission errors caused by the inherent stochasticity of wireless channels. One notable impact of wireless channels on FL is the potential bias introduced towards users with favorable channel conditions. This bias stems from the location-dependent successful decoding for the model updates at BS. In particular, the probability of successfully decoding the transmission of a model update depends on the Signal-to-Interference-Noise Ratio (${\rm SINR}$) of the transmitted signals, which is a function of the user location and channel condition. Operating of a low communication rate that ensures high transmission reliability for cell-edge devices would lead to slow convergence of the global model. In hostile environments,  the assumptions of well-intentioned and accurate model updates do not hold. In real-life scenarios, it is not realistic to assume the absence of cyberattacks or defective devices among the participating clients. As a result, the accuracy and reliability of the learned model could be compromised.

Motivated by the importance and challenges of wireless FL, the authors in \cite{Salehi, Ruslan, r3} engineered a multiplicative factor to counterbalance the bias effect of FedAvg. Furthermore, the authors in \cite{r3, Over_the_Air} developed approaches to accelerate the training process of wireless FL. However, the studies in \cite{Salehi, Ruslan, r3, Over_the_Air} overlook the potential security threats. Such security threats are more prominent for accelerated FL algorithms \cite{r3}, where the presence of compromised devices will significantly impact the global model accuracy. This is due to the scarce data sources in accelerated FL models as the number of participating devices in the training process varies based on the ${\rm SINR}$ conditions \cite{r3}. A security aware FL model is proposed in \cite{trust, COLD}, which involves a trustworthiness raking for devices based on their prior participation history. However, the work in  \cite{trust, COLD} overlooks the wireless communication challenges that include global model bias and slow convergence rate. To the best knowledge of the authors, this paper is the first to account for the combined impact of trustworthiness and wireless network characteristics on FL.

This paper develops a risk-aware FL aggregation model that accounts for wireless communication factors (e.g., interference and fading) as well as the trustworthiness of each device. The main contributions of this paper can be summarized as follows,
\begin{itemize}
\item Developing a risk-aware framework that considers both wireless impairments and security aspects. 
\item Investigating the impact of non-trustworthy users on accelerated wireless FL with scarce data sources. 
\end{itemize}
The results show that being agnostic to the users trustworthiness score leads to significant model deterioration. On the other, relying on fully trusted users is not sufficient due to the data scarcity. The developed model strikes a unique balance between accuracy and convergence speed by including users of moderate trustworthiness score in the initial training rounds and excluding them at the fine-tuning state.

\section{System Model and Problem Formulation}\label{system}
\begin{figure}[t]
\centerline{\includegraphics[width= 8cm]{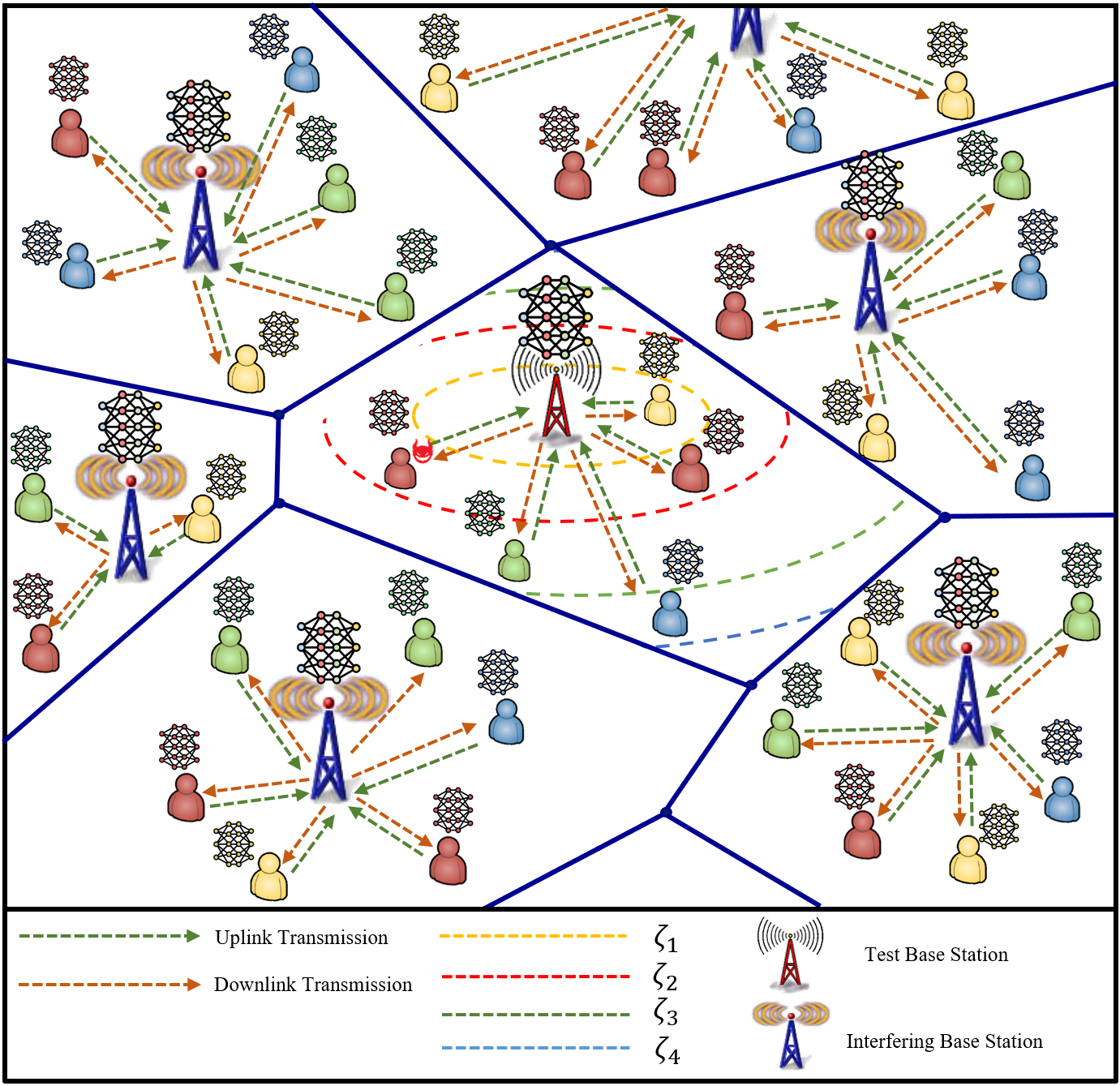}}
\caption{System model showing a test BS implementing dynamic ${\rm SINR}$ based aggregations with 4-levels and five clients in an FL training process.}
\label{Vilinoi Cells with dynamic SINR inclusion}
\end{figure}

Our model considers a single-tier cellular network with BSs located according to a Poisson point process (PPP). The users are assumed to be uniformly distributed in the plane, where each user associates to its geographically closest BS as shown in Fig.~\ref{Vilinoi Cells with dynamic SINR inclusion}. Without loss of generality, we consider a typical BS located at an arbitrary origin and its set of associated users denoted by the set ${U}$.
 
\subsection{Federated Learning Model}
In the set of users ${U}$, each user is indexed by $n$ and has its own dataset $D_n$, which is used for updating the local weights denoted as $\boldsymbol{w}_{n,t}$. The local parameters are updates as follows,
\begin{equation}\label{SGD}
\boldsymbol{\boldsymbol{\psi}}_{n,t}^{(i+1)} = \boldsymbol{\psi}_{n,t}^{(i)}-\gamma g_n(\boldsymbol{\psi}_{n,t}^{(i)})
\end{equation}
where $\boldsymbol{\psi}_{n,t}^{(i)}$ represents the model's parameters for user $n$ in the $i^{th}$ iteration at round $t$. The parameters are updated according to the stochastic gradient descent (SGD) algorithm with a step size $\gamma$, where $g_n(\cdot)$ denotes the gradient of the objective function. The iteration number $i$ ranges from 1 to the total number of local epochs $E$, representing the number of passes over the training data during the optimization process. At the final iteration of the local epochs, $\boldsymbol{\psi}_{n,t}^{(E-1)}$ represents the local weights $\boldsymbol{w}_{n,t}$ that will be transmitted to the BS. The BS then aggregates the successfully received weights using FedAvg to generate the global weight denoted as $\boldsymbol{g}_t$, where $D$ is the entire dataset calculated as $D={\sum_{n=1}^U {D}_{n}}$. A pictorial illustration of the system model is shown in Fig. 1.

During the updating process, the server combines the distributed optimization task as follows,

\begin{equation}\label{minimize}
\min_{\boldsymbol{w}} f(\mathbf{g}) = \frac{1}{U} \sum_{n=1}^U f_n(\boldsymbol{w}),
\end{equation}
where $f_n(\boldsymbol{w})$ represents the loss function associated with user $n$. It is also assumed that the BS has a validation set that is used to track the accuracy of the global model with the training iterations.

\subsection{Communication Model}
The considered network adopts universal frequency reuse scheme for a set of orthogonal channels such that each user $n$ occupies one resource block (RB). Following this approach and having the number of RBs greater than or equal to the number of devices in each cell would allow the users to send their data simultaneously without interfering with each other. However, interference is generated from devices occupying the same RB at different BSs.
To simplify the analysis, we assume perfect downlink communication. In addition, we adopted constant power transmission $P$ for all users in ${U}$. We assume that the noise inherent in the BS, denoted by $N_0$, follows a Gaussian distribution with variance $\sigma^2$, and the small-scale fading, denoted by $h_0$ follows a Rayleigh distribution. Moreover, we adopted the power-law path-loss model wherein the signal strength decays as $r^{-\eta}$ with the increase in propagation distance $r$, where $\eta$ represents the path-loss exponent. 

For accelerated FL convergence, we allow users with high data rates during the initial training rounds. The transmission rate requirement is then relaxed in subsequent rounds to include more users with the progressing training round. Hence, the transmission rate in the $t^{th} $ round is set to $\log(1+\zeta_t)$, where the transmission of the $n^{th}$ user is successful if and only if ${\rm SINR}_{n}^{(t)} > \zeta_t$. A pictorial illustration of the employed dynamic ${\rm SINR}$ training process is shown in Fig. \ref{Vilinoi Cells with dynamic SINR inclusion}. 

\subsection{Trustworthiness Model}
The users are ranked with a trustworthiness score obtained from their profiles, type of devices, and participation history in prior FL processes. The trustworthiness score follows the dual-faceted approach that encompasses two key factors:

\begin{itemize}
\item \textbf{Quality:} This metric provides insights into the data quality as well as the computing capabilities and accuracy of the devices. 
\item \textbf{Security:} This metric provides insights into the user's authenticity and security profile (e.g., security software and updates). 

\end{itemize}
The synthesis of the quality and security metrics leads to a unified measure referred to as users’ trustworthiness metric \cite{trust} \cite{trust2}. This comprehensive metric provides an assessment of users that is crucial for the FL model integrity. To account for random users profiles in large scale networks, the trustworthiness score $\Omega_n \sim Beta(\alpha, \beta)$ is modeled using the beta distribution with parameters $\alpha$ and $\beta$. The choice of the beta distribution is driven by its adaptability and flexibility as a continuous distribution that varies from 0 to 1, allowing for nuanced representation and adaptation to the dynamic nature of the trustworthiness score. 

Users that have $\Omega_n \geq \rho$ are categorized as fully trusted users $U_{F} \subset U$, where $\rho$ is a high threshold that distinguishes reliable devices with up-to-date security patches. On the contrary, users with a score of $\Omega_n \leq \kappa$ are considered malicious users $U_{M} \subset U$, where $\kappa$ is a low threshold that distinguishes faulty devices and adversary users. Users with a score of $ \rho <\Omega_n < \kappa$ are risky users $U_{R} \subset U$ that are reporting the following manipulated weights  
\begin{equation}
    \boldsymbol{w}_{n,t}' = \boldsymbol{w}_{n,t}*\left(1+\frac{(1-\Omega_n)}{10}\right),
    \label{attack_eq}
\end{equation}
which can be due to low-quality data, imprecise computation at the device, or a sort of covert (i.e., hard to detect) model poisoning attack \cite{poison}. In all cases, we assume that the magnitude of the model deviation is inversely proportional to the trustworthiness score of the user $\Omega_n$. This modeling approach enables us to incorporate the impact of users with varying trust scores, effectively capturing the impact of risky users. By employing this equation, we simulate the manipulative actions of risky users and account for model imperfections from users with moderate trust scores. 

It is worth noting that each user in $U$ falls in only of the trustworthiness categories such that $U_{F} \cup U_{R} \cup\ U_{M} =U $ and $U_{F} \cap U_{R} \cap\ U_{M} =\phi$. While $U_{F}$ are fully trusted, they do not possess sufficient data for FL training. Hence, we propose utilizing the set of users $U_{R}$ in the initial training stages. Then, restrict $U_{F}$ for fine-tuning the model during the later stages of the learning process.

\section{Risk-Aware Accelerated Wireless Federated Learning with Heterogeneous Clients}

\subsection{Success Uploading Model}

As discussed earlier, the uplink transmission of the local weights at rate $\log(1+\zeta_t)$ is considered successful if the ${\rm SINR}$ for the transmission exceeds $\zeta_t$. This condition ensures that the received signal strength is sufficient for reliable communication between the client and the BS. If the received power falls below this threshold, the quality of the received signal is insufficient for successful data decoding. Excluding users who fail to meet the trustworthiness score requirements would introduce an imbalance in contributions to the global model's training process. Thus, rectifying this potential imbalance by assigning a location-dependent weight, which amplifies the contribution of farther devices due to their less contribution in the communication rounds.

\begin{lemma}\label{weighted-factor-lemma}
The location-dependent weighted factor denoted by $S_{n,t}(\zeta_t,r_n)$ for a randomly selected client $n$ is given by,
\end{lemma}
\begin{equation}
    \begin{split}
        \frac{1}{S_{n,t}(\zeta_t,r_n)}&=\frac{\exp\left(\frac{\zeta_t N_0}{P r_n^{-\eta}} \right)}{\mathcal{L} \left( \frac{\zeta_t}{P r_n^{-\eta}}\right)},
    \end{split}
\end{equation}

\begin{IEEEproof}
See Appendix A.
\end{IEEEproof}

As a result, as shown in Lemma \ref{weighted-factor-lemma}, for a randomly selected user the model will be weighted by $\frac{1}{S_{n,t}(\zeta_t,r_n)}= \frac{1}{\mathbb{P}({\rm SINR}_n^{(t)} > \zeta_t)}$. The weighting factor consists of the probability that the device's ${\rm SINR}$ exceeds the threshold $\zeta_t$. It operates as an amplification mechanism to the contribution of each user to the global model, such that devices with lower success probability have higher contributions. Here $\mathcal{L}(s)$ represents the Laplace transform of the interference in the uplink, which can be obtained via systematic stochastic geometry analysis as \cite{Stochastic}:

    \begin{align}
    \mathcal{L}(s) 
    = \exp\left(-2\pi\lambda \int_{0}^{\infty} \left( \frac{(1-\exp{(-\pi \lambda r^2)})}{1+\frac{r^\eta}{sP}}\right) r dr \right).
    \label{lablace}
    \end{align}

\subsection{Dynamic ${\rm SINR}$ Thresholds and Trustworthiness Integration: Algorithm Details}
\textbf{Algorithm \ref{alg:dynamicSINR}} outlines the enhanced dynamic ${\rm SINR}$ approach with an incorporation of trustworthiness considerations. 
It is often assumed that a limited dataset is available, typically derived from publicly accessible datasets or provided by the service provider. This practice aims to ensure similarity in data distribution to the private data held by clients, as demonstrated in prior works such as \cite{bsdatabase}\cite{bsdatabase2}. In the initial rounds, the system accommodates the set of users $U'$ that contains both fully trusted and risky users $U_{F} \cup U_{R}$. This inclusion of users is particularly effective in the context of non-IID datasets, where diverse devices with distinct datasets contribute positively to system enhancement. Upon receiving the local weights by the BS, the algorithm proceeds by categorizing these users into a newly defined subset, denoted as $U"$. This subset only contains users who have met the ${\rm SINR}$ requirements. 
\begin{equation}
    \begin{split}\label{updateequation}
        \boldsymbol{g_{t+1}} \gets \boldsymbol{g_{t}}+ \frac{1}{U'} \sum_{n=1}^{U'}\frac{\mathbbm{1}{\{{\rm SINR}_n^{(t)}>\zeta_t\}}}{S_{n,t}(\zeta_t,r_n)}(\boldsymbol{w}_{n,t}-\boldsymbol{g_{t}}),
    \end{split}
\end{equation}
where $\mathbbm{1}\{\cdot\}$ is the indicator function which takes the value 1 when $\{\cdot\}$ is true and zero otherwise. As the communication rounds progress, the updating formula (\ref{updateequation}), which accounts for the cumulative contributions of eligible users, leads to an accumulation of noise arising from the assumptions in the adversary model. This cumulative effect reaches a critical point, detrimentally impacting the overall model. At this point, including $U_{R}$ no longer contributes positively to the system; on the contrary, it degrades the performance. Consequently, the system dynamically transitions to exclusively considering $U_{F}$ to optimize and fine-tune overall performance. The dynamic transition can be achieved by utilizing a trust window $\mu$. The system only considers authenticated users if the global accuracy evaluated at the BS decreases within the trusted window $\mu$.

\begin{algorithm}
\SetAlgoLined
\KwData{$\zeta_t,\theta_n,\boldsymbol{w}_0,E,T,\gamma,\mu$}
\KwResult{$\boldsymbol{g}_t$}
\label{alg:dynamicSINR}
Initialization\;
$\boldsymbol{g}_0 \gets \text{initial value depends on the learning task}$\;
$\boldsymbol{S} \gets \text{array of size } T$\;  
$U' \gets U_{F} \cup U_{R}$\; 
\For{$t \gets 0$ \KwTo $T-1$}{
    \For{$n \gets 1$ \KwTo $U'$ \text{\qquad\qquad In Parallel} \\}{
    $\boldsymbol{\boldsymbol{\psi}_{n,t}^{(0)}} \gets \boldsymbol{g_{t}}$\\
    
    \For{$i \gets 1$ \KwTo $E$}{
    $\boldsymbol{\boldsymbol{\psi}_{n,t}^{(i)}} \gets \boldsymbol{\boldsymbol{\psi}_{n,t}^{(i-1)}} -\gamma \mathbf{g_n}(\boldsymbol{\boldsymbol{\psi}_{n,t}^{(i-1)}}) $
}
$\boldsymbol{w}_{n,t} \gets \boldsymbol{\boldsymbol{\psi}_{n,t}^{(i)}} $\\
transmit   $\boldsymbol{w}_{n,t}$
}
 \If{${\rm SINR}_n^{(t)} > \zeta_t$}{
            Add user $n$ to $U''$\;
        }
 $\boldsymbol{g_{t}} \gets \boldsymbol{g_{t-1}}+ \frac{1}{U''} \sum_{n=1}^{U''}\frac{1}{S_{n,t}}(\boldsymbol{w}_{n,t}-\boldsymbol{g_{t-1}})$\\
$\boldsymbol{S}[h] \gets $ evaluate the accuracy at ($\boldsymbol{g_{t+1}}$)\;
\If{$\boldsymbol{S}[h]$ is smaller than the $\mu$ preceding elements}{
$U' \gets U_{F}$\;

}
}

\Return Result\;
\caption{Risk-Aware Accelerated Wireless Federated Learning}
\end{algorithm}
 \section{Numerical Results}
We consider the density of the BS $\lambda$ to be $50/km^2$, and the simulation area is 3000 x 3000 $km^2$. Each BS has 30 RBs; hence, it can serve up to 30 devices. The path loss exponent $\eta$ is set to 4 for the urban environment, and the transmission power for all devices is set to 10 dBm. Regarding the machine learning model, we use a 2-layer convolutional neural network followed by two fully connected layers while having the optimizer set to be a momentum stochastic gradient descent (SGD) with one epoch, a momentum of 0.5, and a learning rate $\gamma$ of 0.01. We trained the model on the MINSIT digits classification dataset \cite{MINSIT}. Concerning the modeling parameter of trustworthiness, the coefficients $\alpha$ and $\beta$ are picked differently to achieve different trustworthiness means. For a trustworthiness mean of 0.95, we used 11 and 1, respectively. For a trustworthiness mean of 0.85, we used  5 and 1, respectively. Finally, for a trustworthiness mean of 0.75, we used 3 and 1, respectively. We set $\rho$ and $kappa$ to be 0.9 and 0.3 respectively. The dynamic threshold associated with the ${\rm SINR}$  is established from 10 dB to 0 dB with a 0.25 step size.  

\begin{figure}[h]
\centerline{\includegraphics[width= 9cm]{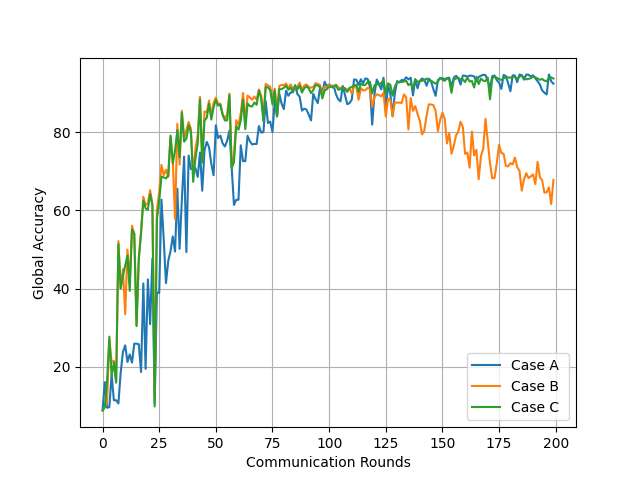}}
\caption{Global Loss vs Communication Rounds with mean = 0.95.}
\label{Global_Loss_High_trust}
\end{figure}

\begin{figure}[h]
\centerline{\includegraphics[width= 9cm]{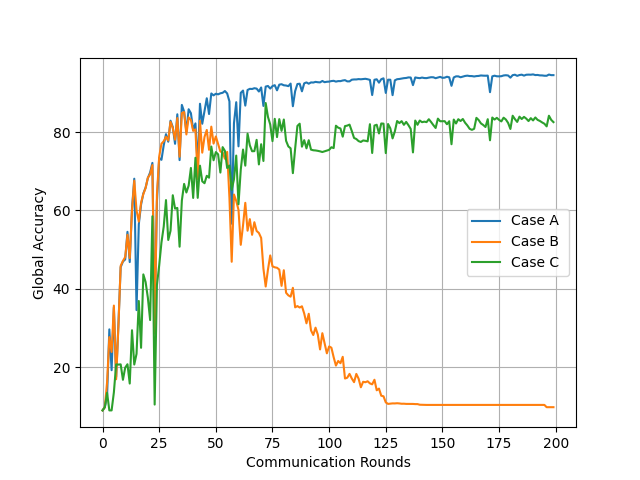}}
\caption{Global Loss vs Communication Rounds with mean = 0.85.}
\label{Global_Loss_Mid_trust}
\end{figure}

\begin{figure}[h]
\centerline{\includegraphics[width= 9cm]{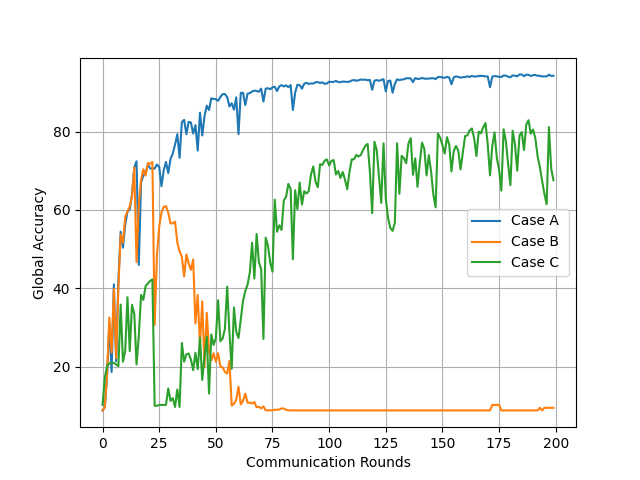}}
\caption{Global Loss vs Communication Rounds with mean = 0.75.}
\label{Global_Loss_Low_trust}
\end{figure}

For the numerical results, we discern the following three distinct cases for comparison. 
\begin{itemize}
    \item \textbf{Case A (Risk-Aware):} all users in $U_F$ and $U_R$ are considered until global accuracy starts degrading and then switching to $U_f$ only for fine-tuning. 
    \item \textbf{Case B (Risk-Agnostic)}: all users in $U_F$ and $U_R$ are considered till the end of the simulation. 
    \item \textbf{Case C (Conservative):} only fully trusted users $U_F$ are considered. 
\end{itemize}

From the numerical results, several insights emerge. First, adopting a conservative approach is insufficient in terms of the convergence rate and the maximum accuracy achieved. This effect is more noticeable in networks with lower trustworthiness, as shown in Fig. \ref{Global_Loss_Mid_trust} and Fig. \ref{Global_Loss_Low_trust}. Second, there is a specific point that further consideration of risky users $U_{R}$ will negatively affect the overall system due to the cumulative added noise for each user. Third, the risk-aware approach, which involves fully trusted and risky users in the early rounds and fine-tuning the model using only the authenticated users results in the highest accuracy and a high convergence rate. The gap between the two approaches widens as the trustworthiness mean gets lower. Another crucial observation lies in the paradoxical outcome that, despite neglecting a portion of the users, there is an observed increase in accuracy. Remarkably, the model does not exhibit bias towards the retained users. This intriguing phenomenon is attributed to the fact that, when certain users are removed from the training process, the remaining users may still possess data corresponding to the same classes as those omitted. This intricacy ensures that the removal of specific users does not compromise the accuracy of the model. Moreover, the amount of data required to fine-tune an existing model is less than the required to construct it.


\section{Conclusion}
In this work, we explored the domain of Federated Learning (FL), a widely recognized distributed machine learning paradigm, we conducted a thorough analysis uncovering the inherent heterogeneity among clients. This heterogeneity manifested in variations related to security, computational resources, and closeness to the base station. To address the challenges posed by this diversity, we developed a dynamic system that strategically operates at a high data rate during the initial stages of communication rounds, thereby effectively speeding up the learning process. As the learning advances, the system adjusts its parameters, aiming to accommodate a broader user base seamlessly. Our experimental findings underscored the significance of a clear proposed approach of being risk-aware. Contrary to the conservative approach of disregarding users with moderate trust scores, we discovered that such exclusion did not necessarily yield enhancements to the system. Likewise, adopting a Risk-Agnostic approach, which involves incorporating all fully trusted and risky users results in significant model degradation in the long run. The pivotal contribution of our work lies in introducing a new algorithm that considers the combined impact of trustworthiness and wireless network impairments. This innovative algorithm combines elements from both conservative and Risk-Agnostic approaches, demonstrating a substantial enhancement in various network trust environments. In conclusion, our study sheds light on the intricate dynamics of FL in the face of client heterogeneity and presents a novel algorithm that represents a breakthrough in optimizing FL for diverse and complex scenarios. 
In future work, a deeper exploration of different attack schemes will be considered, aiming to enhance the comprehensiveness of our analysis. Additionally, there is a need to analyze the optimal transition point from a Risk-Agnostic to a Risk-Aware approach, ensuring a more nuanced and adaptive security strategy.


\section{Appendix}

Lemma 1 can be proved as follows 
\begin{equation}
    \begin{split}
    \frac{1}{S_{n,t}(\zeta_t,r_n)}&= \frac{1}{\mathbb{P}({\rm SINR}_n^{(t)} > \zeta_t)}  \\
    \frac{1}{S_{n,t}(\zeta_t,r_n)}&= \frac{1}{\mathbb{P}\left( \frac{P h_0^2 r^{-\eta}}{N_0 +I_{agg}} > \zeta_t \right)}    \\
    &\overset{(a)}{=} \frac{1}{ \mathbb{E}_{Iagg} \left[\exp\left(\frac{-\zeta_t \left( N_0 + I_{agg}\right)}{P r^{-\eta}}\right)\right]}\\
    &\overset{(b)}{=} \frac{1}{\exp\left(\frac{-\zeta_t N_0}{P r^{-\eta}} \right) \cdot \mathcal{L} \left( \frac{\zeta_t}{P r^{-\eta}}\right)}\\  
    &= \frac{\exp\left(\frac{\zeta_t N_0}{P r^{-\eta}} \right) }{\mathcal{L} \left( \frac{\zeta_t}{P r^{-\eta}}\right)}\\
\end{split} \notag
\end{equation}
where (a) is achieved by the cumulative distribution function (CDF) of the exponential distribution of the fading and (b) is obtained by substituting the expectation on the aggregated interference. Note that $\mathcal{L} \left( s \right)$ is the Laplace transform of the probability density function (PDF) of the aggregated interference. 

\end{document}